\documentclass[sigconf, nonacm]{acmart}
\usepackage{enumitem}
\usepackage{booktabs}
\usepackage{adjustbox}
\usepackage{multirow}
\usepackage{colortbl}
\usepackage{xcolor}         
\usepackage{caption} 

\newcommand{\hide}[1]{} 
\newcommand{\vpara}[1]{\vspace{0.05in}\noindent \textbf{#1 }}
\newcommand{\ipara}[1]{\vspace{0.05in}\noindent \textit{#1 }}

\newcommand{\model}{\textsc{E2ETune}}

\newcommand{\smodel}{\textsc{E2ETune} }

\newcommand\vldbdoi{10.14778/3718057.3718073}
\newcommand\vldbpages{1466 - 1480}
\newcommand\vldbvolume{18}
\newcommand\vldbissue{5}
\newcommand\vldbyear{2025}
\newcommand\vldbauthors{\authors}
\newcommand\vldbtitle{\shorttitle} 
\newcommand\vldbavailabilityurl{https://github.com/RUCKBReasoning/E2ETune}
\newcommand\vldbpagestyle{empty}
\usepackage{framed}

\definecolor{shadecolor}{rgb}{0.9,0.9,0.9}
\title{\model: End-to-End Knob Tuning via Fine-tuned Generative Language Model}

\begin{document}


\author{Xinmei Huang}
\affiliation{%
  \institution{Renmin University of China}
}
\email{huangxinmei@ruc.edu.cn}

\author{Haoyang Li}
\affiliation{%
  \institution{Renmin University of China}
}
\email{lihaoyang.cs@ruc.edu.cn}

\author{Jing Zhang}
\affiliation{%
  \institution{Renmin University of China}
}
\email{zhang-jing@ruc.edu.cn}

\author{Xinxin Zhao}
\affiliation{%
  \institution{Renmin University of China}
}
\email{zhaoxinxin798@ruc.edu.cn}

\author{Zhiming Yao}
\affiliation{%
  \institution{Renmin University of China}
}
\email{yaojimmy2005@ruc.edu.cn}

\author{Yiyan Li}
\affiliation{%
  \institution{Renmin University of China}
}
\email{liyiyan@ruc.edu.cn}

\author{Tieying Zhang}
\affiliation{%
  \institution{ByteDance Inc}
}
\email{tieying.zhang}
\email{@bytedance.com}

\author{Jianjun Chen}
\affiliation{%
  \institution{ByteDance Inc}
}
\email{jianjun.chen}
\email{@bytedance.com}

\author{Hong Chen}
\affiliation{%
  \institution{Renmin University of China}
}
\email{chong@ruc.edu.cn}

\author{Cuiping Li}
\affiliation{%
  \institution{Renmin University of China}
}
\email{licuiping@ruc.edu.cn}

\thanks{Xinmei Huang and Haoyang Li contributed equally to this work. Work done during the internship at ByteDance.}
\thanks{Jing Zhang and Tieying Zhang are the corresponding authors.}

\begin{abstract}
Database knob tuning is a significant challenge for database administrators, as it involves tuning a large number of configuration knobs with continuous or discrete values to achieve optimal database performance. Traditional methods, such as manual tuning or learning-based approaches, typically require numerous workload replays and are both time-consuming and resource-intensive. To address this challenge, we introduce \model, an end-to-end knob tuner powered by a fine-tuned generative language model. The key idea is to leverage the exceptional sequence-to-sequence modeling capabilities of generative language models to capture the complex mapping between workloads (inputs) and their corresponding promising configurations (outputs). To achieve this goal, we propose a novel data generation framework to efficiently produce a large amount of training data, where each data sample consists of a workload and its promising configuration. Then, these data are used to fine-tune a generative language model, yielding an end-to-end knob tuner. This tuner offers out-of-the-box configuration recommendations for new workloads. We conduct extensive experiments to evaluate \model's efficiency and effectiveness using 10 representative and 3 real-world benchmarks. Compared to state-of-the-art methods, \smodel can identify competitive configurations in significantly less time. 

\end{abstract}

\maketitle

\pagestyle{\vldbpagestyle}
\begingroup\small\noindent\raggedright\textbf{PVLDB Reference Format:}\\
\vldbauthors. \vldbtitle. PVLDB, \vldbvolume(\vldbissue): \vldbpages, \vldbyear.\\
\href{https://doi.org/\vldbdoi}{doi:\vldbdoi}
\endgroup
\begingroup
\renewcommand\thefootnote{}\footnote{\noindent
This work is licensed under the Creative Commons BY-NC-ND 4.0 International License. Visit \url{https://creativecommons.org/licenses/by-nc-nd/4.0/} to view a copy of this license. For any use beyond those covered by this license, obtain permission by emailing \href{mailto:info@vldb.org}{info@vldb.org}. Copyright is held by the owner/author(s). Publication rights licensed to the VLDB Endowment. \\
\raggedright Proceedings of the VLDB Endowment, Vol. \vldbvolume, No. \vldbissue\ %
ISSN 2150-8097. \\
\href{https://doi.org/\vldbdoi}{doi:\vldbdoi} \\
}\addtocounter{footnote}{-1}\endgroup

\ifdefempty{\vldbavailabilityurl}{}{
\vspace{.3cm}
\begingroup\small\noindent\raggedright\textbf{PVLDB Artifact Availability:}\\
The source code, data, and/or other artifacts have been made available at \url{https://github.com/RUCKBReasoning/E2ETune}.
\endgroup
}


\section{INTRODUCTION}
Knob tuning in databases involves adjusting the values of various knobs to optimize performance for specific workloads. This process is essential because default database configurations are not always suitable for every workload. Effective knob tuning can significantly improve databases' performance and reliability. However, this task is NP-hard due to the vast number of tunable knobs and their wide configuration ranges. Traditionally, database administrators (DBAs) manually perform knob tuning, relying on their experience—a time-consuming and impractical approach for numerous workloads and database instances. With advances in machine learning, researchers have developed automated knob tuners, primarily including Bayesian Optimization (BO)-based methods (\emph{e.g.}, iTuned~\cite{ituned}, SMAC~\cite{SMAC}) and Reinforcement Learning (RL)-based methods (\emph{e.g.}, CDBTune~\cite{CDBtune}, UDO~\cite{udo}). 

\vpara{Limitations of Existing Methods.}
Although existing BO and RL-based methods are capable of autonomously identifying appropriate configurations, they face challenges with tuning efficiency. In particular, for each new given workload, these methods typically require extensive iterations to achieve the desired database performance. Each iteration involves three steps: exploring a new configuration from a learned model, replaying the workload under the new configuration, and refining the model based on the database feedback. This process often involves numerous workload replays, resulting in suboptimal tuning efficiency. For example, for the TPC-H benchmark~\cite{tpch} at a scale factor of 6, our preliminary studies reveal that HEBO~\cite{hebo}, a robust BO method for hyperparameter optimization, takes approximately 23 hours to identify a promising configuration within 100 iterations. To speed up the tuning process, researchers have proposed various transfer learning methods to reduce the number of required iterations. For example, workload mapping~\cite{aken2017@ottertune}, model ensemble~\cite{restune, onlinetune}, and model pre-training~\cite{Qtune} aim to initialize models using historical tuning data to potentially quicken model convergence speed. Additionally, the knob pruning method~\cite{OpAdviser, gptuner, DBbert} focuses on narrowing the search space of configurations, by pinpointing essential knobs and their optimal ranges, thus improving tuning efficiency. However, even with these methods in place, finding a promising configuration still demands dozens to hundreds of iterations. Moreover, when the new workload significantly differs from those previously encountered, transfer approaches become almost ineffective.

\begin{figure}[t]
    \centering
    \includegraphics[width=0.47\textwidth]{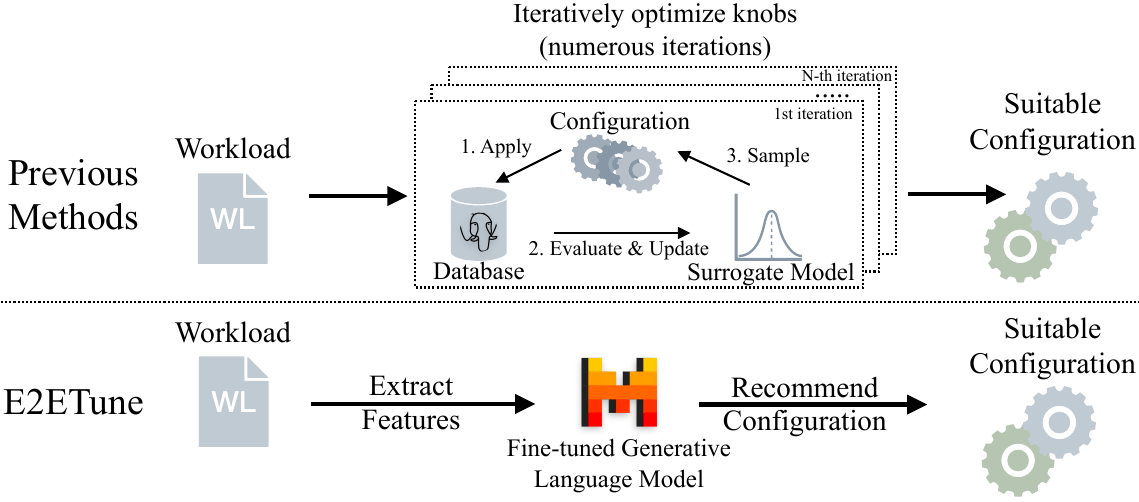}
	\caption{\label{fig:workflow} Previous knob tuning methods \textit{vs.} \model.}
\end{figure}

\vpara{Motivation and Our Proposal.} As described above, the inefficiency of previous methods mainly derives from the iteration process, which incurs a lot of workload replays. To further speed up the tuning process, we propose to develop a novel, end-to-end knob tuning method named \model. Figure~\ref{fig:workflow} illustrates the core concept of \model, which reimagines the knob tuning task as an end-to-end modeling process, effectively eliminating the time-consuming iterative nature of previous methods. To achieve this goal, inspired by the observation that skilled database administrators (DBAs) are often able to manually determine a suitable configuration for a new workload, drawing on their extensive experience, \textit{we posit that there exists a complex distribution-mapping relationship between workloads and their optimal configurations.} Hence, by training a model capable of understanding the distribution-mapping relationship, we can establish an end-to-end solution for knob tuning.

\vpara{Challenges and Our Solutions.} However, developing such an end-to-end knob tuner presents significant technical challenges: (1) In our preliminary attempts, the first challenge we encounter is that traditional machine learning algorithms, such as multilayer perception (MLP) and random forest, can not efficiently learn the distribution mapping from the training data due to their limited learning capabilities. To tackle this challenge, motivated by the powerful end-to-end modeling capabilities of generative language models (LMs)\footnote{In this paper, we specifically use ``language model'' and its abbreviation ``LM'' to denote the sequence-to-sequence generative language model.}, such as GPT-4~\cite{openai2023gpt4} and LLaMA-3~\cite{meta2024@llama3}, fine-tuning a language model to learn and capture the complex distribution mapping will be a feasible solution. Therefore, in this paper, we innovatively formalize the knob tuning task as a sequence-to-sequence generation task, where the input is the features of the workload and the output is the promising knob configuration. (2) Then, to effectively train an end-to-end knob tuner, it's crucial to obtain a substantial dataset of <workload, promising configuration> pairs. Yet, the absence of publicly accessible datasets fulfilling this requirement poses a significant hurdle, marking the scarcity of training data as the second major challenge. To overcome this challenge, we introduce a novel training data construction framework designed to efficiently gather the necessary training data. This framework is structured around two principal components: the generation of workloads and the collection of labels. In the first component, our goal is to generate a large number of diverse and high-quality workloads, leveraging existing database instances. Following this, in the second component, we employ HEBO~\cite{hebo} to identify a promising configuration for each synthesized workload as its label. Furthermore, to expedite the process of label collection, we have developed a cost model that serves as a substitute for actual workload executions within the iterative process of HEBO. In practice, after obtaining enough training data, we fine-tune Mistral-7B~\cite{jiang2023mistral}, a Transformer-based~\cite{vaswani2017attention} generative language model with 7 billion learnable parameters, to perform end-to-end knob tuning.

\vpara{Evaluation.} To thoroughly evaluate \model's effectiveness, we use 10 representative benchmarks—TPC-H, TPC-DS, JOB, SSB, SSB-flat, TPC-C, Twitter, Wikipedia, Smallbank, and YCSB—and 3 real-world benchmarks: StackOverflow, SSAG, and AMPS. As \smodel is fine-tuned using a wide range of diverse and high-quality workloads, it may exhibit the ability to generalize effectively to new test workloads across various database instances, whether seen or unseen during training, without the need for additional fine-tuning. Therefore, our evaluations encompass both ``in-schema'' and ``cross-schema'' scenarios. In the ``in-schema'' scenario, we evaluate \model's performance on test workloads from database instances included in the training sets (for example, training on generated TPC-H workloads and testing on the TPC-H official workload). Conversely, the ``cross-schema'' scenario explores \model's capacity to generalize to test workloads involving database instances it has not encountered previously (for instance, training on generated TPC-DS workloads and testing on the TPC-H official workload).

The main contributions of this paper are as follows:

\begin{itemize}[leftmargin=1em]
\setlength\itemsep{0em}
    \item \textbf{End-to-end Database Knob Tuning:} We introduce \model, a data-driven, end-to-end database knob tuning method based on the fine-tuned language model. It analyzes workload features and then directly recommends promising configurations, eliminating the need for numerous iterations. To the best of our knowledge, \smodel is the first LM-based end-to-end knob tuning method.
    \item \textbf{Novel Data Generation Pipeline:} To gather training data for \model, we develop a novel data generation pipeline. This involves synthesizing new workloads, labeling them with appropriate configurations, and introducing a new cost model as a proxy for actual execution to significantly speed up the data collection process.
    \item \textbf{Comprehensive Evaluation:} Our extensive evaluations, spanning 10 representative and 3 real-world benchmarks, reveal that \smodel not only identifies the most effective configurations but also significantly enhances tuning efficiency, surpassing existing methods. Moreover, detailed ``cross-schema'' experiments demonstrate that \smodel can smoothly adapt to new workloads on previously unseen database instances.
\end{itemize}
\section{RELATED WORK}
\nocite{*}

\subsection{Knob Tuning System}
A knob-tuning system typically comprises a knob tuner to identify promising configurations and a knowledge transfer module to leverage historical knowledge. In contrast to prior studies, \smodel integrates knowledge transfer into the core knob-tuning process, representing a novel approach. We will now review existing knob tuners and knowledge transfer methods.

\subsubsection{Knob Tuning} Existing knob tuning methods can be classified into 4 categories: heuristic-based methods, Bayesian Optimization (BO)-based methods, Reinforcement Learning (RL)-based methods, and Deep Learning (DL)-based methods.

\vpara{Heuristic-based methods.}
Heuristic-based methods involve exploring the search space through manually crafted rules~\cite{dageville2002@sql_memory_management} or a set of predefined heuristic rules~\cite{ansel2014@opentuner}. Nonetheless, these approaches demand significant human intervention and often exhibit limited search efficiency, resulting in suboptimal performance.

\vpara{BO-based methods.}
BO-based methods, such as VBO~\cite{VBO}, HEBO \cite{hebo}, and SMAC \cite{SMAC}, use surrogate models to estimate database performance metrics from given features. Each method uses a different surrogate model: VBO employs a vanilla Gaussian process, while SMAC uses a random forest. These approaches require many iterations to gather tuning observations. In each iteration, a configuration is sampled from the surrogate model and applied to the database. The workload is executed to obtain performance metrics, which refine the surrogate model. This iterative process enables surrogate models to iteratively improve their accuracy, leading to optimized database configurations.

The pioneering knob tuning system iTuned \cite{ituned} uses a GP model as its surrogate. Subsequent systems have enhanced surrogate accuracy by incorporating various features. For instance, OnlineTune \cite{onlinetune} includes query arrival rates, types and indexes. CGPTuner \cite{CGPtuner} and RelM \cite{Relm} focus on system-level attributes like memory control across workloads, containers, and JVM setups. ResTune \cite{restune} adds resource utilization metrics such as CPU, memory, and I/O usage. Despite advancements, BO-based models often require hundreds of iterations to recommend a suitable configuration for a new workload. This process is time-consuming and resource-intensive, highlighting the need for efficiency improvements in tuning.

\vpara{RL-based methods.}
RL-based knob tuning methods \cite{CDBtune, Qtune, WATuning} use reinforcement learning algorithms to adjust database knobs. The Deep Deterministic Policy Gradient (DDPG) \cite{ddpg}, known for its actor-critic framework, is widely used in these systems. The actor selects a configuration based on the state, while the critic evaluates its efficacy, optimizing the policy and value function iteratively. DDPG handles continuous action spaces effectively, making it suitable for knob tuning. CDBTune \cite{CDBtune} employs DDPG, using runtime metrics as the state to generate database configurations. QTune \cite{Qtune} enhances this approach with Double-State DDPG (DS-DDPG), incorporating query-related features, such as the types of queries and the tables involved, to improve performance. However, like BO-based methods, RL-based approaches often require many iterations to achieve stable performance, which can be resource-intensive.

\vpara{DL-based methods.}
DL-based knob tuning methods \cite{DNN, ibtune} train neural networks as learned cost models to predict database performance from given features, facilitating rapid configuration exploration. For instance, \cite{DNN} uses a deep neural network instead of the Gaussian process model in OtterTune \cite{aken2017@ottertune}. Similarly, iBTune \cite{ibtune} trains a deep neural network to estimate response time using metrics like logical reads and CPU usage. However, these learned cost models are usually just part of the tuning system, which may still require iterative trials and adjustments.

\subsubsection{Knowledge Transfer}\label{sec:knowledge_transfer}
Traditional knob tuning methods often require numerous iterations to identify promising configurations, posing efficiency challenges. To address these cold-start issues, researchers have developed various knowledge transfer methods: (1) Workload mapping (OtterTune \cite{aken2017@ottertune}) uses past tuning data to improve BO-based tuners' initial states. (2) Model pre-training (QTune \cite{Qtune}) pre-trains RL models with historical data, then fine-tunes them on new workloads for faster convergence. (3) Model ensemble (ResTune \cite{restune}) combines models from past tasks to adapt to new workloads. (4) Knob pruning (GPtuner \cite{gptuner}, OpAdviser \cite{OpAdviser}) refines the configuration space by selecting critical knobs and their optimal ranges based on past experiences, reducing the search space.

Despite these advancements, base tuners still need additional iterations for new workloads, creating a bottleneck. Our paper proposes a novel approach to streamline the tuning process by modeling it as an end-to-end task using historical data, aiming to eliminate the need for multiple iterations.

\subsection{Language Models for Databases}

Recent research in databases has extensively harnessed LMs to optimize various aspects such as
database diagnosis (\emph{e.g.}, root cause analysis)~\cite{dbgpt, zhou2023@dbot}, text-to-SQL~\cite{li2023resdsql,pourreza2023@dinsql, li2024@codes}, data preparation (\emph{e.g.}, value filling and entity resolution)~\cite{tang2021@rpt}, data integration~\cite{suri2021@ember, arora2023language}, query processing~\cite{trummer2022codexdb}, and table question answering~\cite{herzig2020@tapas, ye2023@dater, jiang2022@omnitab}.

Notably, DB-BERT \cite{DBbert} and GPTuner \cite{gptuner} leverage LMs for knob tuning by using knowledge from web forums and manuals to constrain the search space, but still require a base tuner for iterative exploration.
In contrast, our \smodel trains a language model on extensive historical tuning tasks to directly predict suitable configurations, eliminating the need for online tuning. This model, after offline training, can be applied to various workloads across different database instances, significantly enhancing tuning efficiency.

\section{Problem Definition}
\label{sec:problem}

\begin{figure*}[ht]
    \centering
    \includegraphics[width=0.9\linewidth]{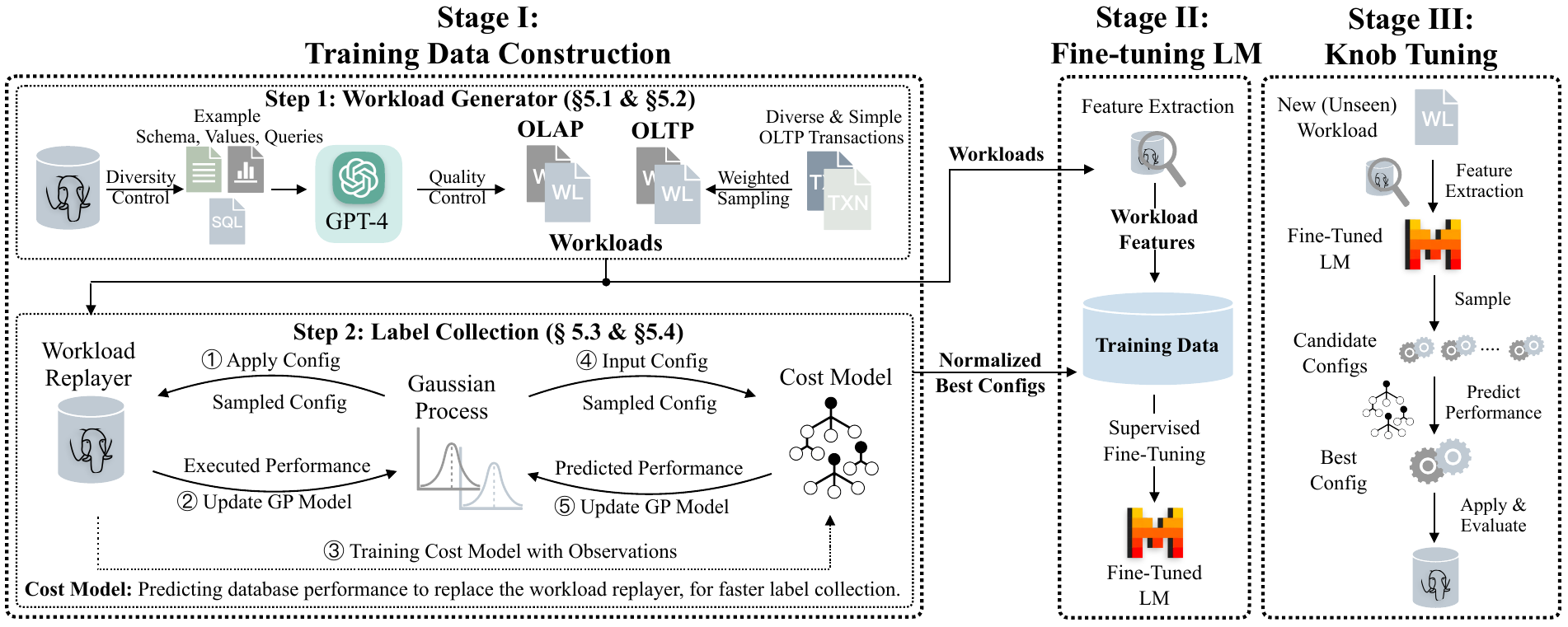}
    \caption{Overview of \smodel (Refer to Section~\ref{sec:overview} for detailed explanation).}
    \label{fig:overview}
\end{figure*}

\textit{Definition 1 (Database Knob Tuning).}
In a database system, there is a set of adjustable knobs, denoted as $K = \{k_1, k_2, ..., k_n\}$. These knobs control various configurable aspects of the database, such as work memory size and maximum connection limits. Each knob $k_i$ can assume a value $s_i$ that falls within a permissible range $S_i$, which can be either continuous numerical values or categorical forms. The entirety of feasible configurations forms a multi-dimensional space for the database system, denoted by $S = S_1 \times S_2 \times ... \times S_n$. A unique configuration within this space is represented by a set of knob values \(\mathbf{s} = (s_1, s_2, ..., s_n) \in S\). Given a workload \(W = \{q_1, ..., q_m\}\) consisting of $m$ queries and a database instance \(D\), the objective of knob tuning is to find the optimal configuration \(\mathbf{s^{*}} = (s_1^{*}, s_2^{*}, ..., s_n^{*})\) within the multi-dimensional space to optimize a performance metric \(M\), such as minimizing latency or maximizing throughput.

\noindent\textit{Definition 2 (End-to-End Knob Tuning via LMs).}
Given a database instance \(D\) and a workload \(W\), \smodel aims to train a language model to learn the distribution mapping function between the workload features and the optimal configuration:
\begin{equation}
    LM_{\Theta}(Feature(D, W)) \rightarrow \mathbf{s^{*}},
\end{equation}
where \(LM_{\Theta}(\cdot)\) denotes the LM, \(\Theta\) represents the learnable parameters of the LM, $Feature(\cdot)$ denotes the function for extracting features of workload $W$ on its corresponding database instance $D$, and \(\mathbf{s^{*}} = (s_1^*, s_2^*, ..., s_n^*)\) represents an optimal configuration that maximizes or minimizes the performance metric \(M\). 

Since this is a data-driven method, it requires a substantial collection of \(<D, W, \mathbf{s^{*}}>\) triplets for training. To obtain such a dataset, we introduce a data generation pipeline detailed in Section~\ref{sec:dataconstruction}. Specifically, we assume the availability of database instances and proceed to automatically generate a diverse array of OLAP and OLTP workloads (refer to Section~\ref{sec:olapworkgeneration} and Section~\ref{sec:oltpworkgeneration}). Subsequently, we capture the configurations associated with these workloads that lead to peak performance metrics (refer to Section~\ref{sec:labelcollection}). After obtaining enough training data, we use them to train a language model in a sequence-to-sequence manner, as outlined in Section~\ref{sec:modeltraining}. Finally, we deploy the trained LM to suggest suitable knob values for new workloads, as elucidated in Section~\ref{sec:knobtuning}.

\section{\smodel Overview}
\label{sec:overview}
The overview of \smodel is illustrated in Figure~\ref{fig:overview}, which consists of 3 main stages.

\vpara{Stage 1: Training Data Construction (Section~\ref{sec:dataconstruction}).}
The first stage involves the offline collection of training data for fine-tuning the LM. This begins by generating diverse OLAP and OLTP workloads across database instances and identifying promising configurations. For OLAP, GPT-4-Turbo creates complex SQL queries based on database characteristics, with diversity and quality control integrated. For OLTP, new workloads are generated by randomly weighting predefined transactions. A BO-based iterative tuner, HEBO, then determines promising configurations to serve as training labels. A cost model accelerates the label collection process during HEBO's tuning. Specifically, real workload replaying generates observations to train the cost model, which then serves as a proxy for real executions during tuning.

\vpara{Stage 2: Fine-tuning LM (Section~\ref{sec:modeltraining}).}
The second stage involves offline fine-tuning of the LM with the data collected in the first stage, teaching it to generate promising configurations based on the features of the workload. More precisely, we gather workload features from three dimensions: workload statistics, query plans, and internal database metrics. Furthermore, to address the language model's limitations in understanding numerical values, we discretize the values of each configuration knob and task the LM with generating the discretized configurations.

\vpara{Stage 3: Knob Tuning with Fine-tuned LM (Section~\ref{sec:knobtuning}).}
Once the LM is trained, it can be deployed to recommend configurations for any given workload. In practice, we sample multiple configurations from the LM by setting the temperature to 1.0 and then we use the cost model trained in stage 1 to select the configuration with the best estimated performance as the recommended result.

\section{Training Data Construction}
\label{sec:dataconstruction}


This training data construction process begins with generating diverse, high-quality workloads tailored to specific database instances. Workloads are categorized into OLAP and OLTP. OLAP workloads focus on complex analytical queries, primarily using \texttt{SELECT} statements for operations like aggregation and slicing. Their performance is measured by latency, the time taken to execute queries. OLTP workloads handle real-time transactional processes, comprising \texttt{SELECT}, \texttt{INSERT}, \texttt{UPDATE}, and \texttt{DELETE} statements. Their efficiency is evaluated by throughput, measured in transactions per second. Given their distinct characteristics, we employ tailored workload generation strategies for OLAP and OLTP.

Subsequently, to identify appropriate configurations for the newly generated workloads, we leverage HEBO~\cite{hebo}, a powerful BO-based knob tuner, to acquire the training labels of the LM (refer to Section~\ref{sec:labelcollection}). Nevertheless, the tuning process of HEBO necessitates numerous workload replaying, significantly impeding the pace of the data preparation. To address this challenge, we introduce a cost model to replace real executions in HEBO's tuning process. This model reduces execution time and allows simultaneous tuning of multiple workloads on a single machine, eliminating resource contention. By integrating HEBO with the cost model, we efficiently generate a large volume of high-quality <workload, suitable configuration> pairs within a feasible offline timeframe.



\subsection{OLAP Workload Generation}\label{sec:olapworkgeneration}
Given that each OLAP benchmark usually comprises a limited number of query templates with variable slots, generating new OLAP workloads by simply substituting these slots with different values might lead to insufficient diversity, potentially undermining the overall generalizability of the fine-tuned language model. Therefore, to create diverse and high-quality OLAP workloads, we first utilize GPT-4-Turbo~\cite{openai2023gpt4, openai2024@gpt4-turbo} to generate realistic and complex SQL queries based on the database information. After obtaining a sufficient number of queries, new OLAP workloads can be obtained by choosing a subset of these queries. 

The prompt utilized to guide GPT-4 to generate OLAP queries comprises six components:
\begin{itemize}[leftmargin=1em]
\setlength\itemsep{0em}
    \item ``Task Overview'' clarifies the objectives for GPT-4-Turbo.
    \item ``Database Schema'' outlines the database instance's structure using Data Definition Language (DDL) statements, including table names, column names, primary and foreign keys, etc.
    \item ``Guidance for Query Generation'' sets constraints for query generation, such as the required SQL dialect.
    \item ``Predicate Generation Aid'' offers some column values to assist GPT-4-Turbo in formulating query predicates.
    \item ``Sample OLAP Queries'' presents a few benchmarked SQL queries to exemplify the desired complexity level of the generated queries.
    \item ``Output Format'' specifies the format of GPT-4-Turbo's output text, aiding in parsing the generated query.
\end{itemize}

\vpara{Diversity Control:} To ensure query diversity, we have implemented several strategies: (1) First, we randomly select tables from the database and incorporate their DDL statements into the ``Database Schema'' part. This approach also reduces API costs associated with invoking GPT-4-Turbo, especially for database schemas with many tables and columns, as the model charges based on the number of tokens it must process. (2) Subsequently, we randomly choose column values from the selected tables to form the ``Predicate Generation Aid'' part.
(3) Finally, we randomly select queries from the benchmarked queries to compose the ``Sample OLAP Queries'' part. To prevent direct copying, we clarify in the prompt that these queries are for illustrative purposes only.

\vpara{Quality Control:} To prevent syntax errors, we use the \texttt{EXPLAIN} command to evaluate the syntactical accuracy of generated queries. If errors are detected, GPT-4-Turbo is prompted to fix them based on the \texttt{EXPLAIN} feedback. Queries with persistent syntax errors are discarded. Among the error-free queries, we execute each one and exclude any that exceed a specified execution time threshold (e.g., over one hour) to mitigate long runtimes.

\subsection{OLTP Workload Generation}\label{sec:oltpworkgeneration}
In the realm of OLTP benchmarks, each benchmark typically comprises a set of predefined transactions. For instance, in the case of TPC-C, five distinct transactions are defined: delivery, new order, order status, payment, and stock level. In contrast to OLAP benchmarks, we observe that adjusting the weights of individual transactions in OLTP benchmarks can result in a wide range of diverse OLTP workloads. This is because the attributes of an OLTP workload, such as the read-write ratio, can be altered by varying the transaction weights. Hence, by manipulating these weights, we can effectively generate a multitude of new OLTP workloads.

\subsection{Label Collection}
\label{sec:labelcollection}
We identify suitable configurations for generated workloads to serve as training labels for the LM using HEBO~\cite{hebo}. Specifically, we perform HEBO to explore the desired configurations. However, HEBO, being a BO-based method, necessitates numerous iterations during the tuning process. Practically, tuning the knobs for a workload demands one to several hours, and a single machine can only tune one workload at a time to avoid resource preemption, significantly slowing data collection.

To address this issue, we introduce a novel cost model to substitute real executions in the iterations. The cost model is trained to predict database performance under a given configurations and workloads. By utilizing this model, we guide HEBO's search process without extensive workload runs. Additionally, it allows concurrent tuning of multiple workloads, further improving the efficiency. Further details regarding the cost model are elaborated in Section~\ref{sec:costmodel}. 


Given that the cost model's predictions may not be entirely precise, therefore, after obtaining the final configuration from the cost model-driven HEBO, we apply it to the database and execute the workload to verify that it surpasses the default configuration in terms of the database performance metric.

\subsection{Cost Model}
\label{sec:costmodel}

\vpara{Cost Model Architecture.}
Research conducted by~\cite{zhang2022facilitating} has shown that Gradient Boosting Regressor (GBR) and Random Forest Regressor (RFR) stand out as powerful regression models, well-suited for the cost estimation task. GBR and RFR are trained independently on the same dataset and their predictions are averaged in an ensemble during inference, reducing overfitting and improving generalization to unseen data.


\vpara{Cost Model Input.}
Given that the core aim of the cost model is to estimate performance metrics under specific configurations and workloads, the model's input comprises two key components: the configuration specifics and workload features. 

To enhance the efficiency, stability, and performance of the cost model, normalization of each knob value in a configuration is crucial due to the varying permissible value ranges across different knobs. We use min-max normalization for each numerical knob value, expressed as:
\begin{equation}
\label{ali:norm-cost-input}
\hat{s}_i = \frac{s_i - \min (S_i)}{\max (S_i) - \min (S_i)},    
\end{equation}
\noindent where \(s_i\) represents the value of the \(i\)-th knob, and \(\max (S_i)\) and \(\min (S_i)\) denote the maximum and minimum values of the \(i\)-th knob, respectively. The range of knob values is typically dictated by hardware specifications, such as memory size, number of CPU cores, etc. In instances where a knob lacks a specified range \(S_i\), a default range from 0 to \(2^{31} - 1\) is assigned. Consequently, \(\mathbf{\hat{s}} = (\hat{s}_1, \hat{s}_2, \ldots, \hat{s}_n)\) denotes a standardized representation of the configuration.

For workload features, we integrate database engine and operating system statistics to characterize the workload comprehensively. Following a careful selection process, we have identified 14 key metrics: xact\_commit, xact\_rollback, blks\_read, blks\_hit, tup\_returned, tup\_fetched, tup\_inserted, conflicts, tup\_updated, tup\_deleted, disk\_read\_count, disk\_write\_count, disk\_read\_bytes, and disk\_write\_bytes\footnote{These metrics are derived from PostgreSQL, which serves as our experimental platform. It's worth noting, however, that other database engines, such as MySQL, offer capabilities to access similar internal metrics.}. The metrics derived from running the workload using the default configuration offer valuable insights for recommending a promising configuration. For example, a high disk\_read\_count indicates that performance bottlenecks may stem from demanding disk read operations, suggesting that adjusting memory-related knobs could potentially improve overall performance.

Similar to configurations, we apply min-max normalization to the metric values based on the allowable range of the metrics to ensure consistency and numerical stability.

Finally, the input to the cost model is the concatenation of the normalized configuration features, denoted as $\mathbf{\hat{s}}$, and the normalized workload features, denoted as $\mathbf{\hat{f}}$. This combined input, $\text{Concat}(\mathbf{\hat{s}}, \mathbf{\hat{f}})$, ensures the cost model incorporates both configuration details and workload characteristics to provide robust and generalized performance estimations. Additionally, since the input vector length remains constant, the trained cost model can be applied to new workloads and database instances.

\vpara{Cost Model Output.}
Given the varying scales of performance metrics across different workloads, directly predicting specific metric values presents significant challenges for the cost model. However, it's crucial to understand that the cost model doesn't need to precisely estimate these values across different workloads since comparing performance metrics between different workloads isn't required. Instead, the primary focus should be on distinguishing between superior and inferior configurations within the same workload, rather than accurately predicting absolute performance metric values. This approach simplifies the learning complexity for the cost model. To achieve this, we normalize the performance metrics across diverse configurations within a specific workload similarly to Eq.~\ref{ali:norm-cost-input}, obtaining \(\hat{p}_{ij}\), the normalized performance metric for the \(j\)-th configuration of the \(i\)-th workload. Finally, the cost model is trained to predict normalized performance based on normalized configuration and workload features.

\vpara{Cost Model Training Data.} To train the cost model, we initially collect a significant amount of tuning observations (\emph{i.e.}, <workload, configuration, performance metric> triplets) as training data using HEBO under the guidance of actual executions. Specifically, we start by randomly selecting 13 newly generated workloads from each of the 10 considered database instances, resulting in a total of 130 workloads. Subsequently, we conduct knob tuning using HEBO alongside real executions, setting the maximum number of iterations to 100, to find promising configurations for the selected workloads. At each iteration of the HEBO process, we capture a tuning observation as a training data sample for the cost model, accumulating a total of 13,000 training data samples. Following this, we train the cost model within a 10-fold cross-validation framework to assess its estimation accuracy. Further evaluation details of the cost model can be found in Section~\ref{sec:quality_of_labels}. Once a dependable cost model is established, we integrate it into HEBO to replace real executions, thereby expediting the label collection process.

\section{Fine-tuning LM}
\label{sec:modeltraining}
Once a sufficient number of historical tuning tasks have been acquired, we can proceed to train a language model to learn the complex relationship between workload and its suitable configuration. This section elaborates on how we formalize the knob tuning task in an end-to-end manner using the LM.

\subsection{LM Input Sequence}
\label{sec:LM-input}
The input to the LM primarily consists of workload features essential for enabling the model to understand and interpret the workload, thereby aiding in generating appropriate configurations. When presented with a workload, we comprehensively assess three dimensions of workload features: workload statistics (information at the workload level), query plans (information at the query level), and internal metrics (information at the system level), all of which are then fed into the LM. Workload statistics offer crucial insights into the inherent characteristics of the workload. Query plans furnish more detailed information regarding the individual queries within the workload. Internal metrics provide granular details about the workload at a system level. Note the workload features input to the LM are much richer than those used for the cost model because the LM can accept a more flexible format and longer input than traditional small GBR and RFR models used for building the cost model. The specific details are outlined below:

\begin{itemize}[leftmargin=1em]
\setlength\itemsep{0em}
\item \textbf{Workload Statistics.}
Workload statistics include: the access frequency of each table, the total number of SQL statements, the read-write ratio, the average number of predicates per SQL query, and the proportion of key operators such as \texttt{ORDER BY}, \texttt{GROUP BY}, and aggregation functions.

\begin{table*}[t]
    \centering
    \caption{Data storage size and the number of training samples in each database instance.}
    \label{tab:benchmark}
    \begin{adjustbox}{width=\textwidth}
    \scriptsize
    \setlength{\tabcolsep}{10pt}
    \renewcommand{\arraystretch}{0.8}
    \begin{tabular}{lcccccccccc}
       \toprule
       Benchmark &  TPC-H~\cite{tpch}  &  TPC-DS~\cite{tpcds} & SSB~\cite{ssb}  & SSB\_flat~\cite{ssb}  & JOB~\cite{Job}  &  TPC-C~\cite{oltpbench}  &  YCSB~\cite{oltpbench}  &  SmallBank~\cite{oltpbench}  &  Twitter~\cite{oltpbench}  &  WikiPedia~\cite{oltpbench} \\
       \midrule
       Size  &  8.6 GB  &  2.1 GB  &  13.3 GB  &  11.7 GB  &  6.9 GB  &  5.0 GB  &  4.0 GB  &  20.0 GB  &  0.4 GB  &  4.4 GB\\
       \#Workload  & 271 & 285      & 299      & 299       & 299      & 300       & 300      & 300      & 300      & 300 \\
  
    \bottomrule
  
    \end{tabular}
    \end{adjustbox}
\end{table*}

\item \textbf{Query Plans.}
Inspired by Qtune~\cite{Qtune}, we integrate the query plans associated with all SQL statements in the workload. The query plans can be obtained using the \texttt{EXPLAIN} command. Unlike prior research that typically utilizes TCNN~\cite{marcus2021@bao} or QueryFormer~\cite{zhao2022@queryformer} for embedding query plans, we treat the query plans as text. Given that the query plan is structured as a sequence, we use nested parentheses to represent the hierarchical relationships between operations. Each pair of parentheses encloses either a single operation or a group of operations executed within a larger step in the plan. Additionally, we augment each operation with the cost estimated by the database engine.

\item \textbf{Internal Metrics.}
Internal database metrics offer crucial insights into workload execution, resource utilization, and potential bottlenecks or inefficiencies within the database system. We utilize the identical set of 14 metrics as inputs for the cost model (see section~\ref{sec:costmodel}). These metrics are obtained by running the workload under the default configuration. To aid the LM in comprehending large numerical values, we simplify them by representing them in orders of magnitude. For instance, we convert ``83,438,203'' to ``83.4 million''.

\end{itemize}

All these features are flattened and concatenated into a sequence before being fed into the LM. The LM's attention mechanism effectively integrates these multi-dimensional features to facilitate the subsequent configuration generation task.

\subsection{LM Output Sequence}
\label{sec:LM-output}
Utilizing the extracted workload features, the LM is trained to model the distribution of suitable configurations. However, generating numerical knob values directly poses challenges due to varying value scales across different knobs. To address this, we normalize the knob values in the configuration using the method detailed in Eq.~\ref{ali:norm-cost-input}. Then, we discretize these normalized values into pre-defined buckets. Specifically, we segment the knob values into 10 buckets, labeled as ``0\% to 10\%'', ``10\% to 20\%'', and so on. This approach enables the LM to generate the buckets corresponding to each knob, which is believed to be easier for the LM to learn compared to outputting exact numerical values. For additional training details, please refer to Section~\ref{sec:impl_details}.

\subsection{Loss Function}
Given an input sequence $x$ and its corresponding output sequence $y$, we optimize the parameters of the LM with a conditional language modeling loss: $p_{\theta}(y|x) = \prod_{t=1}^{l} p_{\theta}(y_t|x, y_{<t}),$ where $\theta$ denotes the learnable parameters of the LM, $l$ denotes the length of the output sequence, and $y_{<t}$ represents all preceding tokens before position $t$ in the output sequence. The primary goal of fine-tuning is to enhance LM's ability to predict the output sequence with maximum likelihood, given the input sequence.

\section{Knob Tuning with Fine-tuned LM}
\label{sec:knobtuning}
After obtaining the fine-tuned LM, we can perform knob tuning in a \textbf{sampling-then-ranking} inference strategy. 

\vpara{Step1: Sampling.}
Given that the LM is fine-tuned to model the probability distribution of suitable configurations based on the workload features. Therefore, we can conduct sampling on the LM's output distribution to obtain multiple candidate configurations during the inference phase. This sampling-based approach significantly boosts the diversity and innovation of the generated configurations, enabling the LM to explore a wider range of potential solutions and identify the optimal configuration. In practice, we typically use a temperature of $1.0$ and set the sampling count to 8. Once the LM predicts a discretized configuration, we convert the buckets into numerical values via denormalization. For instance, if the LM predicts that the ``max\_wal\_senders'' knob falls within the ``30\% to 40\%'' bucket, we calculate the midpoint of this bucket, (30\% + 40\%) / 2 = 35\%, and then obtain the specific knob value using the maximum and minimum values of that knob.

\vpara{Step2: Ranking.}
Following the generation of 8 candidate configurations, we rank them to identify the best option. For this ranking phase, we leverage the cost model introduced in Section~\ref{sec:costmodel} to evaluate and order the sampled configurations, ultimately selecting the best one as the final recommended configuration.

\section{EXPERIMENT}

\subsection{Experimental Settings}

\subsubsection{Datasets.} \label{sec:datasets}

To evaluate \model's performance, we use 10 widely adopted benchmarks, including five OLAP benchmarks (TPC-H, JOB, SSB, SSB-flat, TPC-DS) and five OLTP benchmarks (TPC-C, Smallbank, YCSB, Twitter, Wikipedia). Each benchmark includes a database instance and predefined templates for SQL queries (OLAP) or transactions (OLTP).

We train the model on the synthetic training dataset and test its performance on the original workload of each benchmark. For OLAP's test set, queries are generated by populating templates with random values. For OLTP's test set, default transaction weights are used to create testing workloads.
For the training set, we use the method introduced in Section~\ref{sec:dataconstruction} to generate 300 new workloads per database instance and gather their corresponding labels (\emph{i.e.}, promising configurations). After filtering out underperforming configurations, we have 2,953 training samples for the LM.

The training data originates from two sources: HEBO with real execution and HEBO with a cost model. Real execution tunes 130 workloads for training the cost model. Labels for the remaining 2,823 samples are generated via the cost model. We ensure that original workloads are not included in the training data.
Table~\ref{tab:benchmark} shows data size and training sample count per database instance. 

To further evaluate the practical application value of \model, we additionally consider three benchmarks from real-world scenarios: StackOverflow (4.5GB)\footnote{The database is sourced from https://stackoverflow.blog and benchmarked queries are sources from https://data.stackexchange.com/stackoverflow/queries.}, SSAG (58GB), and AMPS (25GB). StackOverflow is an OLAP benchmark designed to analyze the website data, for example, mining the features of unanswered questions. SSAG and AMPS are two private business workloads in ByteDance. Specifically, SSAG is an OLAP benchmark utilized in slow SQL analysis and governance scenarios, encompassing complex queries such as slow SQL template analysis, logical database analysis, and new slow SQL identification. AMPS, on the other hand, is an OLTP benchmark used in the AI platform services, which includes transactions related to user management, permission control, algorithm management, model management, and task scheduling. These benchmarks are employed solely as test sets.

\subsubsection{Baselines.}

\vpara{Traditional Methods.} We evaluate the BO-based methods SMAC \cite{SMAC} and HEBO \cite{hebo}. SMAC has shown strong performance in database knob tuning \cite{zhang2022facilitating}, while HEBO has demonstrated effectiveness across hyperparameter optimization tasks \cite{hebo}. In addition, we try an RL-based method CDBTune \cite{CDBtune}, which uses the DDPG algorithm for database tuning. All methods are executed for a maximum of 100 iterations, with each iteration involving time-consuming workload replay. Furthermore, we utilize our trained cost model to replace real execution during the iterative process, resulting in three new baselines: SMAC + Cost Model, HEBO + Cost Model, and CDBTune + Cost Model. Finally, we randomly sample 1,000,000 configurations using the Latin Hypercube Sampling strategy from the knob space and employ the cost model to identify the best configuration. This baseline is referred to as Random Sampling + Cost Model.

\vpara{Knowledge Transfer Methods.} 
As categorized in Section~\ref{sec:knowledge_transfer}, we consider four types of transfer methods: workload mapping, model ensemble, model pre-training, and knob pruning.

\ipara{Workload Mapping}: Introduced in OtterTune \cite{aken2017@ottertune}, this technique leverages historical tuning data from similar workloads to initialize a new tuning model. This technique is often integrated into BO-based methods, and we integrate it with HEBO and SMAC.

\ipara{Model Ensemble}: Proposed by Restune \cite{restune}, this technique combines multiple historical knob-tuning models and incorporates workload features. For a new workload, similar historical workloads' models are ensembled. Commonly, this integrates with BO-based methods.

\ipara{Model Pre-training}: QTune \cite{Qtune} introduces this technique, the actor and critic models are pre-trained with past tuning tasks and fine-tuned with RL-based methods. In our experiments, we assess it with CDBTune in two scenarios: with and without online tuning, where online tuning involves fine-tuning based on real-time feedback.

\ipara{Knob Pruning}: This technique reduces the search space. OpAdviser \cite{OpAdviser} refines the search space based on similar historical workloads' insights. DB-BERT \cite{DBbert} and GPTuner \cite{gptuner} leverage language models to extract useful information from database manuals, effectively trimming the search space during tuning.

To ensure fairness, the historical tuning tasks utilized in these knowledge transfer techniques are the training dataset for \model. The model for selecting tuning methods in OpAdviser is trained using their publicly available dataset.

\begin{figure*}[ht]
    \centering
    \captionsetup{
        labelfont={bf},
    }
    \includegraphics[width=1.0\linewidth]{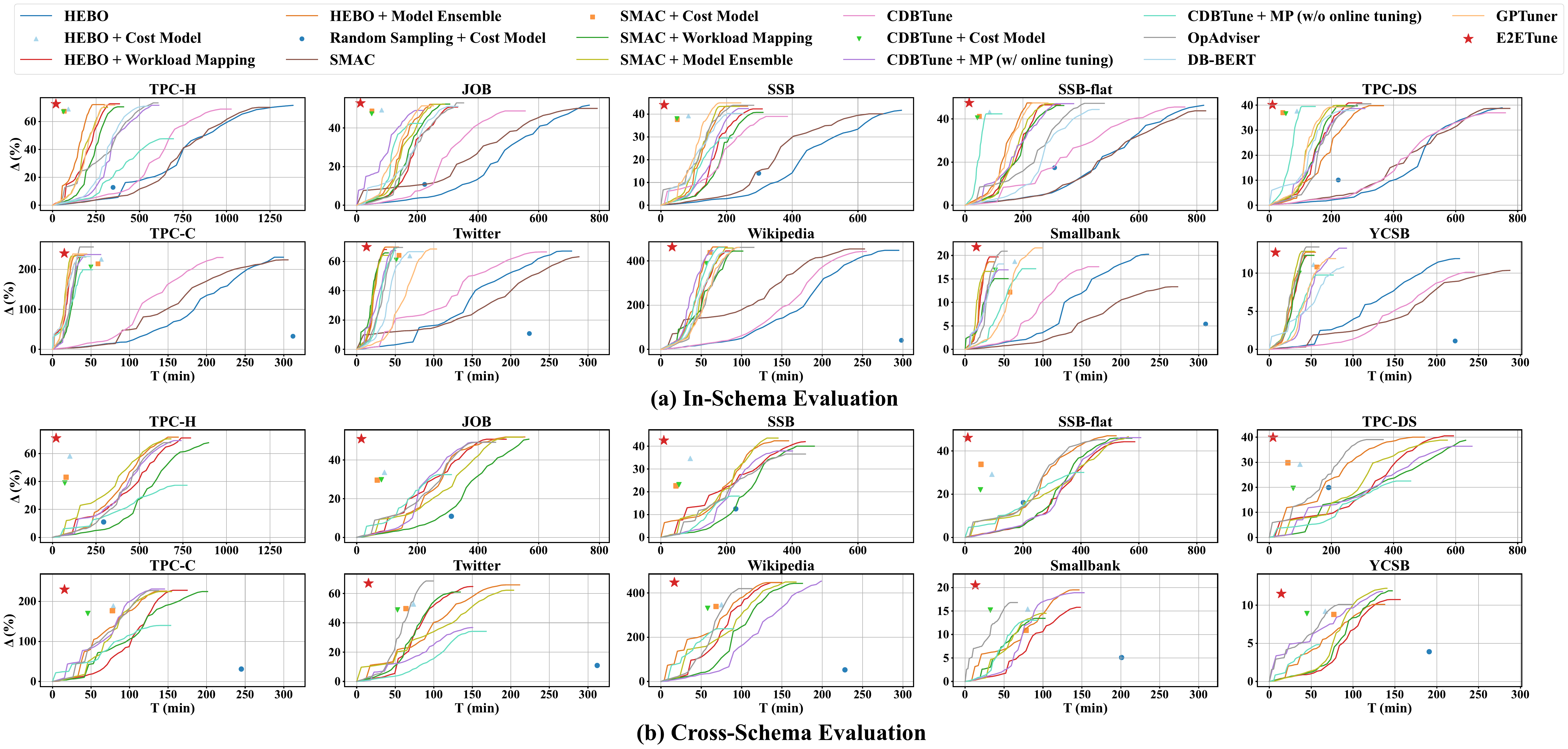}
    \caption{Best performance improvement over tuning time across 10 representative benchmarks. ``MP'' is the abbreviation of Model Pre-training. (top-left is better)}
    \label{fig:main-result}
\end{figure*}

\subsubsection{Metrics.}

For OLAP workloads, our goal is to minimize the query latency. Therefore, the performance improvement is defined as: $\Delta = \frac{\text{default latency} - \text{optimized latency}}{\text{default latency}}$, where ``default latency'' is from the default configuration and ``optimized latency'' is from the recommended configuration. For OLTP workloads, our goal is to maximize throughput (\emph{i.e.}, transactions per second, tps). The performance improvement is defined as $\Delta = \frac{\text{optimized tps} - \text{default tps}}{\text{default tps}}$, where ``default tps'' is from the default configuration and ``optimized tps'' is from the recommended configuration.

\subsubsection{Implementation details.}\label{sec:impl_details}
We use PostgreSQL 12.2 to manage all databases. Following prior studies~\cite{zhang2022facilitating, CDBtune}, we manually select 45 crucial knobs for tuning, while retaining default values for the rest. The DBMS is restarted with each new configuration since some knob modifications require it. For training the LM, we use Mistral-7B-Instruct-v0.2\footnote{\url{https://huggingface.co/mistralai/Mistral-7B-Instruct-v0.2}} as the base model, fine-tuned with PyTorch 2.1~\cite{ansel2024@pytorch2} and Hugging Face Transformers~\cite{wolf2019huggingface}. The learning rate is set to 2e-5, batch size to 128, and maximum context length to 8,192. The model is trained for 4 epochs using a cosine decay learning rate scheduler. To save GPU memory, we use FlashAttention-2~\cite{dao2023@flashattn2} and DeepSpeed ZERO~\cite{rajbhandari2020@zero} for optimized data parallelism.

\subsubsection{Environments.} 
We use a CPU server with an Intel Xeon E5-2650 v4 (12 cores, 24 threads) and 64GB RAM to host the PostgreSQL database for experiments. Fine-tuning and deploying the LM are done on a machine with an Intel Xeon Gold 5218 CPU, 256GB RAM, and 4 NVIDIA GeForce RTX 3090 GPUs. The process of fine-tuning the LM takes approximately 10 hours. These machines are connected via high-speed local networks.

\subsubsection{Evaluation Settings.} \label{sec:evaluation_settings}
The evaluation is conducted in three different settings: \textbf{In-Schema}: The LM is trained on our generated workloads and tested on the original workloads of 10 representative benchmarks. \textbf{Cross-Schema}: We employ a 5-fold cross-validation approach to simulate cross-schema scenarios. In each fold, we select one OLAP and one OLTP benchmark as the hold-out test set, training the LM on our generated samples from the remaining 8 benchmarks. This ensures there is no schema overlap between the training and testing sets, allowing us to evaluate \model's performance on new database instances. \textbf{Real-World}: The LM is trained using our generated data samples and directly applied to recommend configurations for three new real-world benchmarks.

\subsection{Main Results}
\label{sec:main-res}

\subsubsection{In-Schema Evaluation}
Regarding the in-schema evaluation results shown in Figure~\ref{fig:main-result} (a), we have the following findings:

\textbf{\smodel emerges as the fastest knob tuning method while delivering competitive database performance improvements.} It achieves the shortest tuning time across all ten benchmarks, showcasing the advantages of end-to-end modeling. For instance, in the TPC-H benchmark, HEBO takes 1,381.7 minutes; even with workload mapping, it still requires 328.7 minutes. In contrast, \smodel completes the task in just 19.8 minutes, achieving up to a 98.6\% reduction in time \(\left(\frac{1,381.7 - 19.8}{1,381.7} = 98.6\%\right)\). Regarding performance, \smodel identifies the best configurations in four benchmarks (JOB, SSB-flat, Twitter, Smallbank) and the second-best in five benchmarks (TPC-H, SSB, TPC-DS, TPC-C, Wikipedia).

\textbf{\smodel breaks its upper bound, HEBO.} 
Despite the fact that the training labels (i.e., promising configurations) for \smodel are derived from HEBO, \smodel surpasses HEBO's performance across all benchmarks. This unexpected outcome can be attributed to \model's ``sampling-then-ranking'' inference strategy. The sampling introduces randomness, allowing the trained LM to explore a range of configurations around HEBO's upper bound. This insight suggests that in the future, \smodel could potentially be leveraged to replace HEBO in generating training labels, a concept known as ``self-training'' or ``self-improvement''~\cite{huang2023@self-improve}.

\textbf{\smodel surpasses other knowledge transfer methods that utilize historical tuning knowledge (refer to +Workload Mapping, +Model Ensemble, and +MP).} While traditional knowledge transfer techniques can speed up the tuning process, they still require much longer time for tuning compared to \smodel. These methods still rely on the iterative processes of traditional knob tuners. In contrast, \smodel uses historical tuning data to train an end-to-end knob tuner, removing redundant iterative steps. 

\textbf{\smodel outperforms approaches that also use LMs in knob tuning, such as DB-BERT~\cite{DBbert} and GPTuner~\cite{gptuner}.} While these methods and our \smodel both utilize LMs, they differ in their primary objectives. DB-BERT and GPTuner use LMs to constrain the search space for traditional knob tuners, thus they still necessitate numerous workload replays. In contrast, \smodel trains a language model to directly predict suitable configurations. Consequently, \smodel not only significantly accelerates the tuning process but also achieves comparable or superior performance enhancements compared to these methods.

\textbf{\smodel is faster and better than traditional methods assisted by the cost model (refer to +Cost Model).} 
While our trained cost model can replace time-consuming workload execution, traditional methods still require a lot of additional time to update their models online and recommend new configurations with each iteration, resulting in low efficiency. Furthermore, the cost model can not always accurately reflect database performance, leading traditional methods to identify sub-optimal configurations. Moreover, even with random sampling of 1 million configurations, the best option identified by the cost model often performs poorly, highlighting the vast search space inherent in the knob tuning task.

\subsubsection{Cross-Schema Evaluation}
The cross-schema scenario assesses how well a knowledge transfer method handles workloads from new database instances. Traditional knob tuners and two knowledge transfer methods (GPTuner and DB-BERT) exhibit the same behavior in both cross-schema and in-schema scenarios since they don't use historical tuning tasks to initialize their models. As a result, these baselines are excluded from the cross-schema evaluation.

As illustrated in Figure~\ref{fig:main-result} (b), in the cross-schema setting, \smodel not only achieves the highest tuning efficiency but also identifies configurations that are comparable to, or even surpass, those of state-of-the-art approaches. Notably, \smodel discovers the best configuration for the Smallbank benchmark and the second-best configurations for TPC-H, SSB, TPC-C, and Twitter benchmarks.

An interesting discovery is that existing knowledge transfer methods significantly expedite the tuning process and marginally improve database performance in the in-schema setting, but their effectiveness diminishes in the cross-schema environment. This decline can be attributed to several factors. Firstly, techniques such as workload mapping, model ensemble, and OpAdviser heavily rely on workload matching, which becomes challenging when finding similar workloads across different database instances. Additionally, the limited generalization capability of the model pre-training technique, due to the small size of the actor and critic neural networks, contributes to the reduced speedup and performance improvement in the cross-schema scenario.

In contrast, \smodel consistently outperforms baseline methods. Its tuning speed is unaffected by the familiarity of database instances. Thanks to the generalization capabilities of the LM and high-quality training data, \smodel can effectively recommend good configurations for out-of-distribution workloads.

\begin{figure}[ht]
    \centering
    \captionsetup{
        labelfont={bf},
    }
    \includegraphics[width=1.0\linewidth]{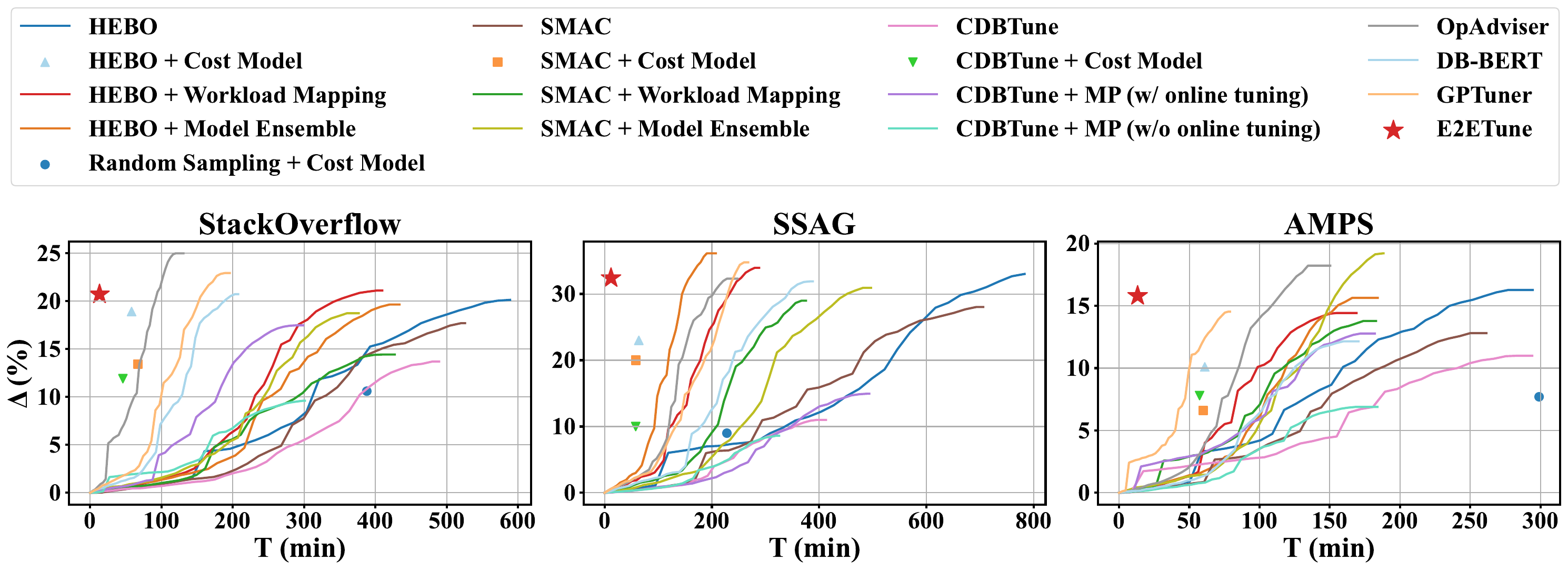}
    \caption{Maximum performance improvement over tuning time on three real-world benchmarks. (top-left is better)}
    \label{fig:main-real}
\end{figure}

\subsubsection{Real-World Evaluation}
\label{sec:main-real-res}
We train \smodel using training samples collected from 10 representative benchmarks and then directly apply it to 3 real-world benchmarks. Notably, for SSAG and AMPS, we implement all baseline methods and our \smodel in ByteDance's internal development environments to assess their effectiveness. These real-world benchmarks introduce several new challenges for \smodel, including more complex queries, larger database scales, private database instances, and different deployment environments.

The experimental results are presented in Figure~\ref{fig:main-real}. In this highly challenging evaluation setting, \smodel exhibits remarkable out-of-the-box capabilities, consistently identifying configurations comparable to those of state-of-the-art methods while significantly reducing the time required. Although \smodel does not achieve the best configurations in this setting, it substantially saves time and resources on tuning tasks, making it more suitable for real-world large-scale deployments.

\subsection{Ablation Study}

\subsubsection{Ablations on Input Features}
As detailed in Section~\ref{sec:LM-input}, the input of \smodel includes workload statistics, query plans, and internal metrics. Ablation studies, shown in Table~\ref{tab:ablation}, reveal performance drops when any component is removed, highlighting their significance. Notably, omitting query plans greatly reduces performance in the cross-schema setting. Query plans are vital as they detail SQL operations and costs, reflecting the database's data distribution, thus aiding the LM in generalizing across diverse database instances.

\begin{table}
    \caption{Ablation studies of \model. 
    We present the average performance improvements $\Delta$ (\%) $\uparrow$  on OLAP and OLTP benchmarks. ``IS'' and ``CS'' represent ``in-schema'' and ``cross-schema'' settings, respectively.}    
    
    \centering
    \begin{adjustbox}{width=0.48\textwidth}
    \setlength{\tabcolsep}{6pt} 
    \renewcommand{\arraystretch}{0.8} 
    \scriptsize
    
    \label{tab:ablation}
    \begin{tabular}{@{}lcccc@{}}
    \toprule
    \textbf{}                & \textbf{OLAP(IS)}  & \textbf{OLTP(IS)}  & \textbf{OLAP(CS)} & \textbf{OLTP(CS)} \\ 
    \midrule
    \model                      & 51.4 & 156.2 & 50.1 & 150.6 \\ \midrule
    Input Features              &  &  &  &  \\
    \quad- w/o internal metrics     & 49.8          & 152.5          & 48.6          & 142.8 \\
    \quad- w/o workload features    & 47.0          & 148.9          & 48.5          & 148.7       \\
    \quad- w/o query plans          & 44.5          & 145.6          & 34.9          & 53.2    \\ \midrule
    LM Inference Strategy      & & & & \\
    \quad- w/o sampling-then-ranking & 50.1          & 152.8          & 47.0          & 144.3 \\  \midrule
    Knob Output Format          &  &  &  &  \\
    \quad- specific values         & 15.1          & 45.3           & 9.4           & 36.8 \\  \midrule
    LM Backbone                   &  &  &  &  \\
    \quad- CodeLLaMA-7B~\cite{roziere2023code}      & 51.3     & 152.7    & 50.5     & 152.8 \\
    \quad- LLaMA2-7B~\cite{touvron2023llama}        & 50.4     & 152.4    & 50.1     & 153.7 \\
    \quad- DeepSeekCoder-7B~\cite{deepseek}      & 48.8     & 159.6    & 48.8     & 135.3 \\
    \quad- DeepSeekCoder-1B~\cite{deepseek}      & 42.0     & 142.4    & 41.0     & 116.1 \\
    \quad- Mistral-7B w/o Pre-training        & NA & NA & NA & NA \\ \midrule
    LM Learning Strategy          &  &  &  &  \\
    \quad- Few-shot Mistral-7B       &  33.8 & 114.2 & 28.6 & 109.5  \\
    \quad- Few-shot GPT-4      &   35.3 &  115.4 & 30.9 & 116.4    \\
    \bottomrule
    
    \end{tabular}
    \end{adjustbox}
    \renewcommand{\arraystretch}{1.0}
    
\end{table}
\subsubsection{Ablations on LM Inference Strategy}
\label{sec:abla-infer}
As outlined in Section~\ref{sec:knobtuning}, we adopt a ``sampling-then-ranking'' strategy to recommend configurations. To evaluate the effectiveness of this inference strategy, we conduct an additional experiment in which we perform inference without utilizing the cost model. Specifically, using the fine-tuned language model, we only recommend a configuration through the greedy decoding method. The evaluation results of this experiment are presented in Table~\ref{tab:ablation} (see ``w/o sampling-then-ranking''). We can see that the ``sampling-then-ranking'' strategy yields slight improvements over the simple greedy decoding method.

\begin{figure}
    \centering
    \includegraphics[width=0.6\linewidth]{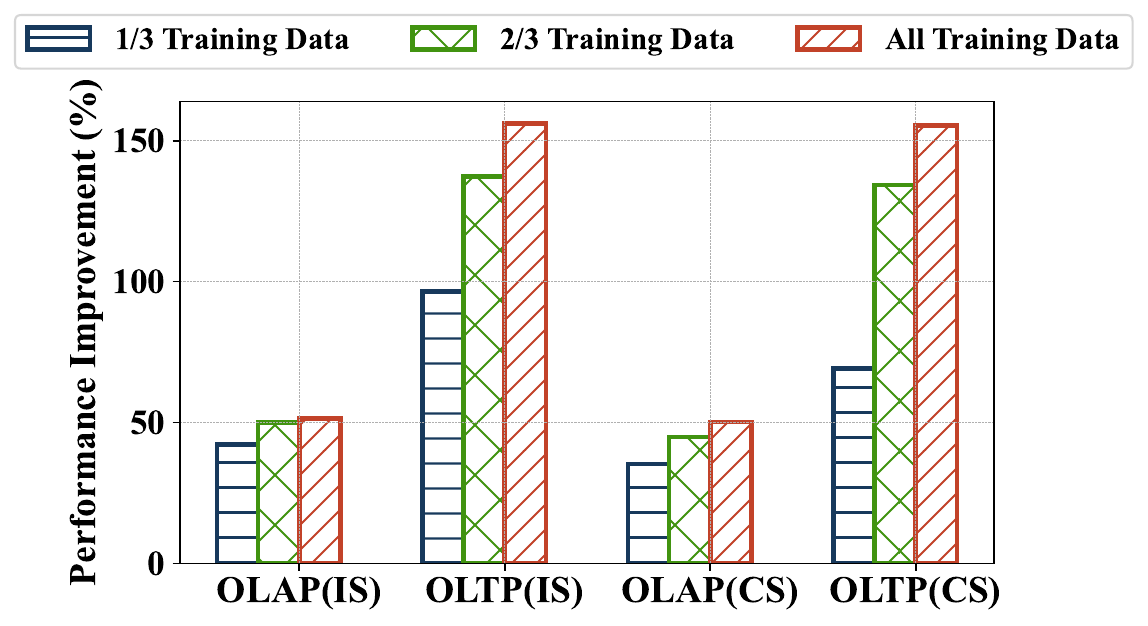}
    \caption{Ablation study of the scale of training data.}
    \label{fig:ablation}
\end{figure}

\subsubsection{Ablations on Output Knob Format}
Section \ref{sec:LM-output}  discusses discretizing knob values into pre-defined buckets. To validate the efficacy of this bucketing approach, we conduct an experiment where the LM directly output knob values (refer to ``specific values'' in Table~\ref{tab:ablation}). The results clearly indicate a significant performance decline when directly outputting values. The poor performance can be attributed to the extensive range and substantial variation in permissible values across different knobs, which poses significant challenges for LMs when tasked with directly outputting knob values. This highlights the effectiveness of the bucketing strategy.

\subsubsection{Ablations on LM Backbone}
We utilize Mistral-7B as the base model of \model. To evaluate the impact of different base models on knob tuning performance, we replace Mistral-7B with other robust models such as CodeLLaMA-7B, LLaMA2-7B, DeepSeekCoder-7B, and DeepSeekCoder-1B to explore model scale effects. Results in Table~\ref{tab:ablation} show that model performance varies by scale and setting: Mistral-7B and DeepSeekCoder-7B excel in OLAP and OLTP benchmarks in the in-schema setting, while CodeLLaMA-7B and LLaMA2-7B perform best in the cross-schema setting. This performance disparity is likely due to differences in pre-training corpora, indicating that a more powerful base model can enhance \model's performance. The 7B version of DeepSeekCoder significantly outperforms the 1B version, highlighting the benefits of larger models.

To explore the impact of the pre-training phase on knob tuning, we test a new Mistral-7B model with randomly initialized weights followed by fine-tuning on our data. This model produce unparseable outputs, underscoring the crucial role of pre-training.

\subsubsection{Ablations on Scale of Training Data}
We delve deeper into the effects of the training data scale on the quality of configurations recommended by the LM. Following the evaluation settings outlined in Section~\ref{sec:evaluation_settings}, we modify the scale of the training dataset for our analysis. Our experiments are conducted across three different scales: 1/3, 2/3, and the complete training dataset, with the results visualized in Figure~\ref{fig:ablation}. The findings illustrate a notable positive relationship between the size of the training data and the enhancement in configuration quality suggested by the LM. It is evident that this trend persists even when utilizing the entire training set, suggesting that the performance of \smodel can be further optimized with an expanded training dataset.

\begin{table*}
    \captionsetup{
        labelfont={bf},
    }
    \caption{A statistical analysis of SQL queries from two sources: those generated by GPT-4-Turbo (denoted as GPT) and those from the benchmark (denoted as Bench). For each statistic, except for the number of distinct templates, the minimum, maximum, and average values are presented as ``MIN / MAX / AVG''.}
    \centering
    \begin{adjustbox}{width=\textwidth}
    \setlength{\tabcolsep}{5pt}
    \scriptsize
    \renewcommand{\arraystretch}{1.0}
    
    \label{tab:query-statistics}
    \begin{tabular}{@{}l|cc|cc|cc|cc|cc@{}}
    \toprule
    \multicolumn{1}{c}{\multirow{2}{*}{Statistics}} & \multicolumn{2}{c}{\textbf{TPC-H}} & \multicolumn{2}{c}{\textbf{JOB}} & \multicolumn{2}{c}{\textbf{TPC-DS}} & \multicolumn{2}{c}{\textbf{SSB}} & \multicolumn{2}{c}{\textbf{SSB-flat}}\\
    \cmidrule(r){2-3}\cmidrule(r){4-5}\cmidrule(r){6-7}\cmidrule(r){8-9}\cmidrule(r){10-11}
    & GPT & Bench & GPT &  Bench& GPT & Bench & GPT & Bench & GPT & Bench\\ \midrule
    \# Referenced Tables per SQL            &1 / 8 / 5.43     & 1 / 8 / 3.41   & 1 / 13 / 5.88  & 4 / 15 / 8.51  &  1 / 11 / 5.46 & 1 / 13 / 5.63   & 2 / 5 / 3.93 & 2 / 5 / 3.76 & 1 / 1 / 1.0 &  1 / 1 / 1.0      \\
    \# Joined Tables per SQL                &{1 / 8 / 5.21}     &{1 / 8 / 2.46}   &{1 / 12 / 5.26}   &{4 / 15 / 8.50}  &{1 / 11 / 5.08} &{1 / 10 / 2.68}    &{1 / 5 / 3.91} &{2 / 5 / 3.76} & {1 / 1 / 1.0} &{1 / 1 / 1.0} \\ 
    \# Referenced Keywords per SQL          &{14 / 34 / 21.96}  &{14 / 29 / 20.27} &{12 / 40 / 26.62}&{11 / 27 / 18.24}& {13 / 42 / 25.41} &{12 / 38 / 23.44} &{14 / 33 / 21.79}&{13 / 19 / 15.77} & {15 / 33 / 22.74} &{12 / 16 / 14.15} \\  
    \# Predicates per SQL                   &{3 / 40 / 15.57}    &{1 / 18 / 6.00}  & {1 / 43 / 16.57} &{14 / 91 / 39.30}  &{3 / 127 / 17.04} &{0 / 96 / 13.05}  & {4 / 46 / 14.33} &{8 / 20 / 13.31} & {1 / 26 / 9.75} & {7 / 16 / 11.31} \\  
    \# Distinct Templates           &{1099}             &{21}             & {1099}           & {99} & {1088}  & {49}  & {1094} &{13}&{1100} & {11}   \\\bottomrule
    \end{tabular}
    \end{adjustbox}
\end{table*}

\subsubsection{Ablations on LM Learning Strategy}\label{sec:in-context-learning}
Besides fine-tuning the LM, another approach is few-shot in-context learning. This method uses a small number of task-specific demonstrations within the LM's context, avoiding the need for extensive fine-tuning. We randomly select 3 samples (<workload features, suitable configuration> pairs) from the training set to use as demonstrations. The original Mistral-7B and GPT-4 models are then instructed to recommend configurations for new workloads using these demonstration prompts. Results in Table~\ref{tab:ablation} (Few-shot Mistral-7B and GPT-4) show that few-shot learning is less effective than full fine-tuning. Even with GPT-4, the tuning performance is below the proposed \model, due to lower demonstration utilization efficiency. 

\subsubsection{Ablations on Modeling Algorithms}
In \model, we utilize a language model to capture the distribution mapping from workload features to their corresponding promising configurations. In this ablation study, we aim to investigate whether traditional machine learning models can also effectively capture this intricate distribution mapping. We evaluate two commonly used machine learning approaches: random forests and multi-layer perceptrons (MLP). To ensure a fair comparison, we maintain consistency in the input and output information with \model. However, some adjustments should be made to accommodate these machine learning techniques. Workload features are represented as a vector that combines workload statistics, query plan features, and internal metrics. Query plan features are extracted using the methodology proposed in~\cite{Qtune}. Then, as we discretize the knob values into buckets for \model, we can reconstruct the output into classification labels for traditional machine learning methods, where each label corresponds to a specific bucket identifier. In the case of the random forest, we specify 1000 trees and a maximum depth of 50. For the multi-layer perceptron, a neural network with three hidden layers, each comprising 128 dimensions, is employed. To mitigate overfitting, batch normalization and dropout layers are incorporated. Given the varying feature lengths across database instances, a separate model needs to be trained for each new instance. Our evaluations are conducted on the TPC-H and TPC-C benchmarks.

Results from the TPC-H benchmark show that the random forest and MLP models yield performance improvements of 36.8\% and 51.3\% respectively, significantly lower than the 72.5\% enhancement achieved by \model. Similarly, on the TPC-C benchmark, the random forest and MLP models exhibit performance gains of 128.8\% and 177.5\% respectively, while \smodel outperforms them with a remarkable 239.9\% improvement. These findings reveal that despite leveraging high-quality input features, traditional machine learning methods struggle to capture the intricate distribution mapping due to their limited modeling and generalization capabilities.

\subsubsection{Ablations on Training Workload Generation Strategy}
\label{sec:abla-template}
In the database field, a common approach to generating workload queries is to use predefined templates~\cite{marcus2021@bao, zhu2023@lero, CEB, yang2022@balsa}. However, we believe this strategy limits the diversity of the generated queries, thereby constraining the generalizability of the trained models. In this section, we conduct an ablation study on the JOB benchmark to support this hypothesis. Specifically, following the Cardinality Estimation Benchmark (CEB)~\cite{CEB}, we employ 13 templates designed for the JOB's database instance, resulting in approximately 1,100 SQL queries combined into 299 distinct workloads. We then collect labels (\emph{i.e.}, promising configurations) for these new template-based workloads using the framework introduced in Section~\ref{sec:labelcollection}. Next, we train two language models using training samples from GPT-4-Turbo-generated workloads and template-generated workloads, denoted as $M_{\text{ours}}$ and $M_{\text{CEB}}$, respectively. Finally, we utilize both $M_{\text{ours}}$ and $M_{\text{CEB}}$ to recommend configurations for the original JOB benchmark workload. The results reveal that the configurations recommended by $M_{\text{ours}}$ yield a 50.26\% performance improvement, while those from $M_{\text{CEB}}$ result in only a 38.56\% improvement. This discrepancy highlights the importance of query diversity in training a highly generalizable language model for knob tuning.

\subsection{Evaluation of Training Data}
The quality of \model's training dataset is crucial. Each training sample consists of a <workload, promising configuration> pair, crafted using the framework outlined in Section~\ref{sec:dataconstruction}. We evaluate the quality of these components individually.

\subsubsection{Quality of Workloads} 
\label{sec:quality_workload}
\textbf{OLAP Workloads.} The OLAP workloads used for training are generated by GPT-4-Turbo. In Table~\ref{tab:query-statistics}, we provide a detailed statistical analysis of both GPT-generated and benchmarked SQL queries across five aspects: the number of referenced tables per SQL query, the number of joined tables per SQL query, the number of referenced keywords per SQL query, the number of predicates per SQL query, and the number of distinct query templates. Our analysis reveals that GPT-generated queries not only effectively cover the benchmarked queries but also significantly enhance the diversity of query templates, underscoring the quality of the queries produced by GPT-4-Turbo. \textbf{OLTP Workloads.} The OLTP workloads are generated by randomly assigning weights to predefined transactions, ensuring high quality and diversity.

\subsubsection{Quality of Labels}\label{sec:quality_of_labels}
In this work, two key factors affect the quality of labels: the knob tuning method and the cost model.

\textbf{Tuning Method.} Our primary objective in choosing a tuning method is to select a powerful and easily deployable knob tuner. In this regard, we assess three potential methods: HEBO, SMAC, and CDBTune. Our experimental findings, detailed in Figure~\ref{fig:main-result}, demonstrate that HEBO surpasses the other two methods in performance. Consequently, we adopt HEBO to gather labels.

\textbf{Cost Model.} The quality of the cost model significantly impacts label quality, as a substantial portion of the training set is derived from it. To provide a comprehensive evaluation of the cost model, we examine two critical aspects: (1) the accuracy of the cost model and (2) the performance of HEBO assisted by the cost model. First, we perform 10-fold cross-validation to train the cost model. For each fold, we use the coefficient of determination ($R^2$) to evaluate the cost model's accuracy on the held-out test set: 
$R^2 = 1 - \frac{\sum_{i=1}^{n} (y_i - \hat{y}_i)^2}{\sum_{i=1}^{n} (y_i - \bar{y})^2}$,
where $y_i$ is the observed performance metric for the $i$-th sample, $\hat{y}_i$ is the predicted performance metric, $\bar{y}$ is the mean of all observed metrics, and $n$ is the total test set observations. Our cost model achieves an average $R^2$ score of 0.862 across the 10-fold validation, indicating high reliability. In addition, during label collection, the cost model is trained using observations from all database instances. Therefore, the quality of the configurations gathered by HEBO with the cost model can be observed in the in-schema part of ``HEBO + Cost Model'' in Figure~\ref{fig:main-result}. Compared with the original HEBO, using the cost model only leads to a marginal decrease in database performance, affirming the reliability of the cost model.

\subsection{Data Collection Cost}
\label{sec:data-collection-time}
To provide clarity on the expenses associated with collecting the complete training dataset, we compute the time for each stage detailed in Section~\ref{sec:dataconstruction}, encompassing both workload generation and label collection times. Importantly, it should be emphasized that the training data collection process can be carried out offline.

\subsubsection{Workload Generation} For OLAP workloads, we employ GPT-4-Turbo to craft approximately 1,100 OLAP queries per database instance. With a  response time of approximately 20 seconds per query and five database instances, the cumulative query generation time amounts to roughly 31 hours, costing approximately \$150.
In contrast, generating OLTP workloads incurs no additional expenses.

\subsubsection{Label Collection}
The training labels of \smodel are sourced from HEBO with real execution and HEBO with the cost model. 

\textbf{Collecting Data via Real Execution.}
For OLAP workloads, executing the full workload takes between 1 to 20 minutes. Since the original HEBO method requires replaying the workload in each iteration, tuning an OLAP workload across 100 iterations may take 2 to 30 hours.
For OLTP workloads, we use a 1-minute stress test for each tuning iteration. As a result, the complete tuning process for an OLTP workload takes approximately 2 hours. 
To ensure accurate test results, workloads are tuned sequentially on the server. With each database instance tuning 13 workloads, the full process takes around 30 days on a single machine. In practice, we use eight identical CPU servers to expedite this process.

\textbf{Collecting Data via Cost Model.}
The results presented in Figure~\ref{fig:main-result} demonstrate that the cost model can significantly enhance tuning efficiency when compared to actual executions. This efficiency boost facilitates the gathering of a substantial volume of training data for \model. Moreover, the elimination of the need for database testing allows for the simultaneous tuning of multiple workloads on a single machine. Leveraging the cost model, despite each database instance needing to tune more than 250 workloads, we can complete this task within 40 days using just one machine. We also utilize eight identical CPU servers to accelerate the process.

\section{DISCUSSION}\label{sec:discussion}

In this section, we first discuss the applicable scenarios for \model. By leveraging the advantages of an end-to-end design, \smodel significantly enhances tuning efficiency while recommending promising configurations. When encountering a new database engine, it is necessary to re-collect training data and train a new model to tune knobs due to different knobs. Consequently, \smodel is particularly well-suited for stable or enumerable environments, such as cloud services that maintain uniform hardware configurations and standardized DBMS products. After the initial training, \smodel proves highly effective in recommending configurations for a diverse range of users utilizing cloud databases. We are deploying \smodel in real-world applications at ByteDance and will open-source our fine-tuned model to support further research. Additionally, to accelerate data collection and enhance label quality, we can explore and leverage better hyperparameter optimization algorithms, such as HPFSO~\cite{hpfso}, SPBOpt~\cite{spbopt}, and DRE~\cite{dre}, as replacements for HEBO in our data collection framework.

\section{CONCLUSION}
This paper explores an innovative method for end-to-end database knob tuning. Our emphasis is on utilizing the advanced language model (LM) to accurately capture the complex mapping between the workload and its promising configuration. To achieve this goal, we introduce a data generation framework to automatically produce workloads and their promising configurations, serving as training data samples for the LM. Through extensive experiments on OLAP and OLTP benchmarks, we demonstrate that \smodel not only significantly improves the tuning efficiency over existing approaches, but also showcases notable database performance improvements in both in-schema and cross-schema settings.

\begin{acks}
This work is supported by the National Key Research \& Development Plan of China (2023YFF0725100) and the National Natural Science Foundation of China (62322214, U23A20299, U24B20144, 62172424, 62276270). We also acknowledge the support of the Public Computing Cloud, Renmin University of China.
\end{acks}

\bibliographystyle{ACM-Reference-Format}
\bibliography{sample}

\end{document}